\documentclass{article}

\usepackage{microtype}
\usepackage{graphicx}
\usepackage{subfigure}
\usepackage{booktabs} 

\usepackage{amsmath}
\usepackage{amssymb}
\usepackage[dvipsnames]{xcolor}
\usepackage{listings}
\usepackage[breakable]{tcolorbox}

\usepackage{hyperref}

\usepackage[accepted]{icml2025}

\usepackage{amsmath}
\usepackage{amssymb}
\usepackage{mathtools}
\usepackage{amsthm}

\usepackage[capitalize,noabbrev]{cleveref}

\usepackage{bbm}
\usepackage{xfrac}
\usepackage{physics}
\usepackage{mdframed}

\theoremstyle{plain}

\theoremstyle{definition}

\theoremstyle{remark}

\usepackage[textsize=tiny]{todonotes}

\icmltitlerunning{Advancing AI-Scientist Understanding: Multi-Agent LLMs with Interpretable Physics Reasoning}

\begin{document}

\twocolumn[
\icmltitle{Advancing AI-Scientist Understanding: \\Multi-Agent LLMs with Interpretable Physics Reasoning}



\icmlsetsymbol{equal}{*}

\begin{icmlauthorlist}
\icmlauthor{Yinggan Xu}{ucla}
\icmlauthor{Hana Kimlee}{nsf}
\icmlauthor{Yijia Xiao}{ucla}
\icmlauthor{Di Luo}{ucla}
\end{icmlauthorlist}
\icmlaffiliation{nsf}{NSF Center for Quantum Network}
\icmlaffiliation{ucla}{University of California, Los Angeles}
\icmlcorrespondingauthor{Di Luo}{\href{mailto:diluo@ucla.edu}{diluo@ucla.edu}}

\icmlkeywords{long context length, language models, information theory, mutual information, large language models (LLMs), machine learning, neural networks, deep learning, generative modeling, genAI, ICML}

\vskip 0.3in
]

\printAffiliationsAndNotice{}
\begin{abstract}
 Large Language Models (LLMs) are playing an increasingly important role in physics research by assisting with symbolic manipulation, numerical computation, and scientific reasoning. However, ensuring the reliability, transparency, and interpretability of their outputs remains a major challenge. In this work, we introduce a novel multi-agent LLM physicist framework that fosters collaboration between AI and human scientists through three key modules: a reasoning module, an interpretation module, and an AI–scientist interaction module. Recognizing that effective physics reasoning demands logical rigor, quantitative accuracy, and alignment with established theoretical models, we propose an interpretation module that employs a team of specialized LLM agents—including summarizers, model builders, visualization tools, and testers—to systematically structure LLM outputs into transparent, physically grounded science models. A case study demonstrates that our approach significantly improves interpretability, enables systematic validation, and enhances human–AI collaboration in physics problem-solving and discovery. 
 Our work bridges free-form LLM reasoning with interpretable, executable models for scientific analysis, enabling more transparent and verifiable AI-augmented research.
\end{abstract}

\section{Introduction}
Large Language Models (LLMs) have become increasingly popular for tackling complex physics problems, emerging as valuable assistants to scientists \cite{zhang2024comprehensive}. However, interpreting the solutions they generate remains a significant challenge due to the inherent complexity of physics problems. Identifying potential flaws often demands substantial effort from experts, as LLM-generated solutions can obscure their underlying reasoning.

Several key issues contribute to this interpretability gap. First, the reasoning trajectories employed by LLMs are often highly complex and diverse. Depending on the inference techniques used, ranging from direct outputs to tool-assisted reasoning, the underlying processes may be partially hidden or require considerable effort to trace. Second, the numerical complexity involved in many physics problems poses a significant verification challenge, making it difficult for humans to independently validate the results. Third, the absence of an interpretable underlying mechanism can lead to seemingly correct outcomes even when the LLM’s understanding of the physics is flawed.

To address these challenges, we develop a novel multi-agent LLM physicists framework that enhances the interpretability, transparency, and verifiability of LLM outputs in physics problem-solving. Unlike prior approaches that treat LLMs as black-box solvers, this framework decomposes the reasoning pipeline into three coordinated modules: a reasoning module, an interpretation module, and an AI-scientist interaction module. We propose an innovative LLM interpretation module, consist of a suite of specialized agents including summarizers, model builders, and testers, which translates opaque LLM outputs into structured, executable, and physically grounded science models. This interpretable interface bridges the gap between AI-generated reasoning and human scientific intuition by supporting validation through code execution, visual inspection, and human-in-the-loop critique. Extensive case studies on textbook-level problems from SciBench demonstrate the framework’s ability to detect flaws, test consistency, and enable interactive validation, thereby offering a new pathway for interpretable, verifiable, and collaborative AI-assisted physics discovery.

\section{Related Works}
\subsection{LLM for Physics} 
Researchers have begun exploring the potential of Large Language Models (LLMs) as reasoning tools in the physics domain \cite{anand2024enhancing, ding2023using, pan2024quantummanybodyphysicscalculations, pang2024physics,wang2023newton}. Studies have demonstrated that LLMs can solve complex word problems requiring calculation and inference, often achieving near human-level accuracy, especially with effective prompting techniques such as few-shot learning using similar examples \cite{ding2023using}, leveraging reinforcement learning from human feedback (RLHF) \cite{anand2024enhancing} or implementing agentic system \cite{pang2024physics}. 

While much of this research focuses on general physics reasoning, recent efforts have applied LLMs to highly specialized domains. Pan et al. \cite{pan2024quantummanybodyphysicscalculations} demonstrated that GPT-4 can perform advanced theoretical derivations, such as deriving Hartree–Fock equations, highlighting LLMs’ potential to automate and accelerate research workflows in theoretical physics. However, as most physics reasonings are complex and domain-specific, existing approaches offer limited support for human scientists to interpret and validate LLM-generated results. The lack of intuitive interfaces for understanding these outputs places a significant cognitive burden on researchers, limiting the practical usability of LLMs in scientific discovery.

\subsection{Verifiable Generation}
A parallel line of research focuses on improving the verifiability and interpretability of LLM outputs. One common approach involves grounding generated content in external sources and providing detailed citations \cite{hennigen2023towards, shen2024citekit, li2024truthreader}. Other methods enhance transparency by generating with more structured and intuitive processes \cite{cecchi-babkin-2024-reportgpt} or enable self-explanatory reasoning \cite{huang2023largelanguagemodelsexplain}.

However, physics reasoning differs fundamentally from tasks based purely on factual retrieval or general logical reasoning. Unlike citation-based fact-checking, physics problem-solving requires structured derivations, adherence to established theoretical frameworks, and quantitative validation. Despite advances in interpretable generation, the challenge of making LLM-generated physics reasoning both understandable and verifiable remains largely unexplored.

\section{System Design}

Building on prior research in LLM-assisted physics reasoning and verifiable AI generation, we propose an interpretation module that enhances both interpretability and validation in physics reasoning. We focus on physics reasoning within the context of problem-solving, which represents its most fundamental form. Our approach employs an agentic system composed of specialized agents, each with a distinct role in structuring the reasoning process. This inference-agnostic pipeline can generate science models for a broad range of problem-solving scenarios, regardless of the implementation of the reasoning module. By explicitly modeling the reasoning process, our system deepens AI-scientist understanding, facilitating more transparent, interpretable, and verifiable AI-augmented scientific reasoning. To clearly articulate our approach, we structure our system into three key modules: a reasoning module, which processes physics problems using naive, tool-using, or agentic LLMs; an interpretation module, which refines AI reasoning into structured science models, executable code, and validation tools; and an AI-scientist interaction module, which facilitates human oversight by enabling experts to analyze, critique, and refine AI-generated reasoning.

\begin{figure*}[ht]
    \centering
    \includegraphics[width=0.9\linewidth]{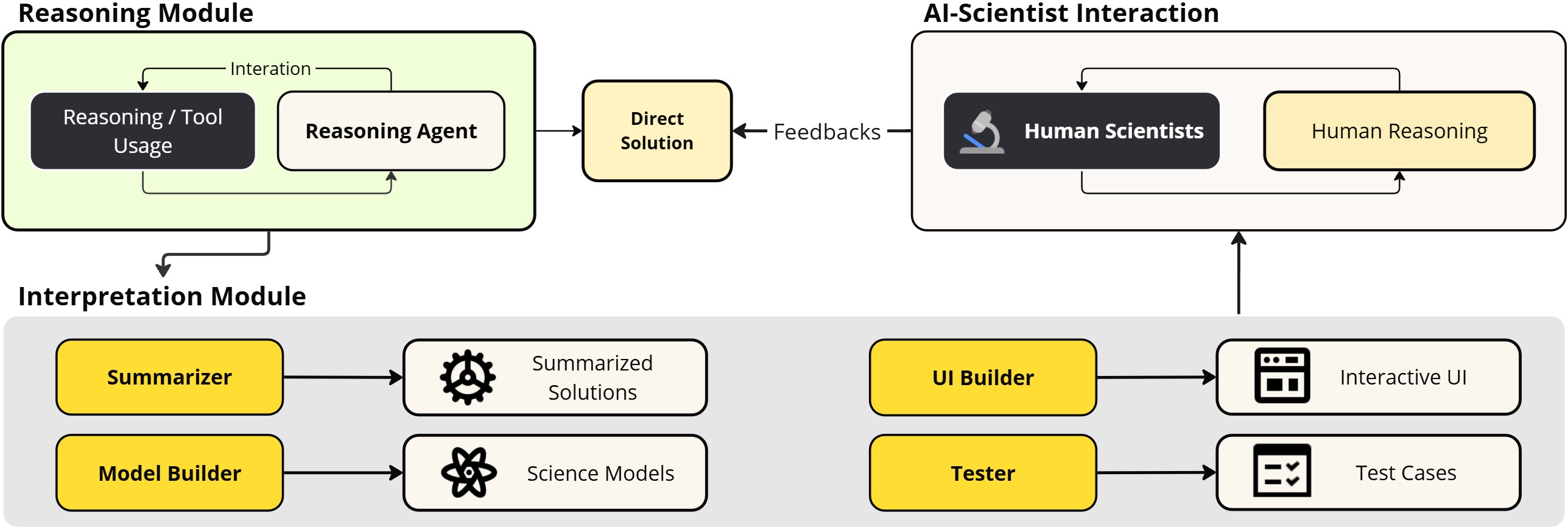}
    \caption{An overview of the augmented reasoning with interpretation module.}
    \label{fig:enter-label}
\end{figure*}

\subsection{LLM Reasoning Module: Establishing the Problem Context}
The reasoning module serves as the entry point to the pipeline, handling diverse physics problems and their solutions from different sources, including: naive LLMs that generate direct, unstructured solutions, tool-using LLMs that incorporate computational resources to refine their responses, and agentic systems that coordinate multiple AI components for enhanced reasoning. While these reasoning modules can be powerful, they often involve complex, opaque processes that may not be fully visible to human scientists. For example, tool-using mechanisms or multi-agent debates can lead to solutions that are difficult to interpret, making it challenging to trace the reasoning behind the results.

\subsection{LLM Interpretation Module: Structuring and Validating AI Reasoning}
To enhance the interpretability and reliability of AI-generated physics solutions, we introduce an interpretation module, which systematically structures AI reasoning into explicit, verifiable science models and provides intuitive feedback for human scientists. Our module refines raw AI outputs into structured representations, aligning them with scientific intuition and enabling validation through interactive tools and automated checks.

This module consists of specialized agents that structure reasoning, build executable models, and enhance human interpretability.

\begin{itemize}
    \item \textbf{LLM Summarizer}
    The summarizer agent processes diverse inputs such as direct solutions, tool usage details, and chat history into a structured, concise format. By preserving core reasoning and reducing redundancy, this agent improves clarity and ensures smoother downstream processing for subsequent agents.

    \item \textbf{LLM Model Builder}  
    To ensure interpretable physics generation, our approach explicitly constructs and validates the underlying science model, which is often implicit in solutions. This module consists of two key components:

        \textbf{Theory Model Builder}: The correctness of an AI-generated physics solution depends on the validity of its underlying conceptual model, which LLMs often leave implicit. This agent explicitly extracts, organizes, and refines the model by identifying key physical quantities, governing equations, and problem constraints. It also uses gater agents classifies the problem type, invokes relevant idealized concepts (e.g. mass point in mechanics) for conceptual coherence.
        
        \textbf{Code Model Builder}: Translating theory models into executable code is essential for validation and downstream applications of the theory model. This agent converts structured science models into computational processes, ensuring consistency between theoretical assumptions and computational implementation.

    \item \textbf{Visualization Builder}
    To support human intuition-driven assessment, the visualization builder generates interactive representations of the coding model. This allows scientists to apply established validation techniques, such as testing extreme conditions and symmetry constraints, to assess solution consistency.

    \item \textbf{LLM Auxiliary Tester}
    While human scientists excel at verification, LLMs can assist this process by performing automated sanity checks like extreme case analysis, providing an additional layer of quality control. Though not a substitute for human judgment, this agent enhances the reliability of AI-generated solutions by identifying inconsistencies.
\end{itemize}

By structuring AI reasoning into explicit science models, executable simulations, and interactive validation tools, the interpretation module improves interpretability, verifiability, and alignment with scientific reasoning.

\subsection{AI-Scientist Interaction: Fostering Collaborative Reasoning}  
Ultimately, our system is designed to augment—not replace—human scientific reasoning. The AI-scientist interaction module ensures that human experts remain central to the validation and refinement process by providing multiple touchpoints for engagement. Scientists can examine and verify the science model to explicitly assess AI reasoning, interact with the visualization interface to dynamically explore and test solutions, and critique AI-generated logic through intuitive representations. By fostering an interpretable reasoning process, this module ensures that AI remains an assistive tool that enhances scientific inquiry while preserving human oversight and expertise.

\section{Case Study and Experiments}

\begin{figure*}[ht]
    \centering
    \includegraphics[width=0.9\linewidth]{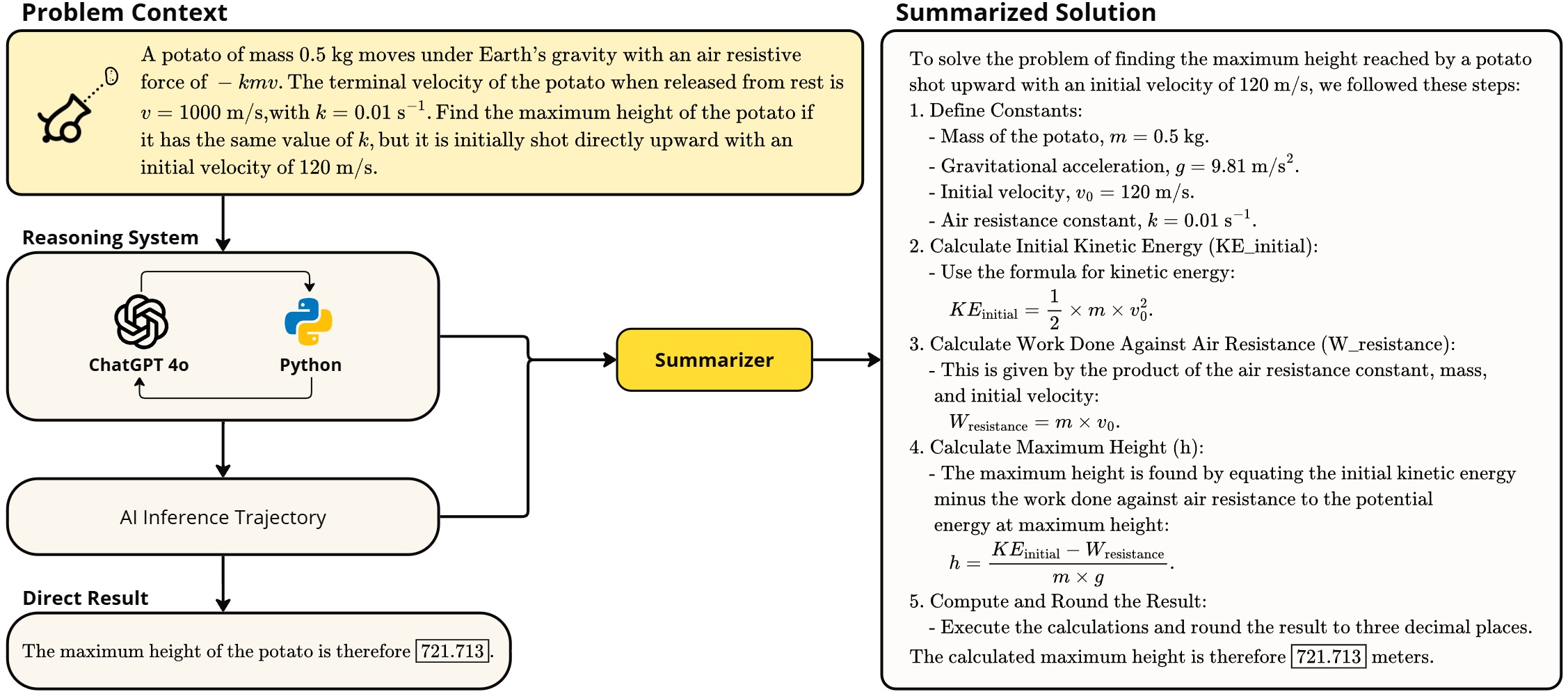}
    \caption{Transformation of a directly generated solution into a summarized solution}
    \label{fig:case-1}
\end{figure*}

We demonstrate the effectiveness of our interpretation module using a mechanics problem from SciBench \cite{wang2023scibench}. In this case, a potato is launched from a potato gun with air resistance, and the task requires an LLM to analyze the object’s motion via the energy conservation law. For our experiments, we utilize ChatGPT-4o \cite{achiam2023gpt} integrated with a Python programming tool as the reasoning module. We use the same prompt templates in SciBench for our reasoning module to solve this problem.

\subsection{LLM Reasoning Module and Summarizer}

Our workflow begins by refining the generated solution through a summarization step. The original inference trajectory includes complex details, including multiple code executions and internal thought processes, which can be difficult for human experts to interpret. Although the direct solution appears to be straightforward, its opaque derivation limits transparency and hinders scientific understanding by human . Our summarizer condenses both the final output and the inference trajectory into a structured form (see Fig. \ref{fig:case-1}). It distills the reasoning trajectory into a step-by-step format for improved interpretability.

\subsection{Model Construction}

Given a problem context and its summarized solution, the interpretation module constructs a corresponding science model in Python and generates an interactive user interface (UI) for scientists to inspect and validate the solution. 

The theoretical model is aligned with the fundamental physics principles familiar to human scientists and serves as a reference for downstream model construction. The Python-based model enables reproduction of numerical results and facilitates modifications to test alternative conditions. The code model follows a predefined template to ensure consistency and a structured format for interpretation and execution. The built models are sent to the downstream agents for testing and user interface construction. We provide full demonstrations and more case studies in the appendix.

\begin{figure*}[ht]
    \centering
    \includegraphics[width=0.6\linewidth]{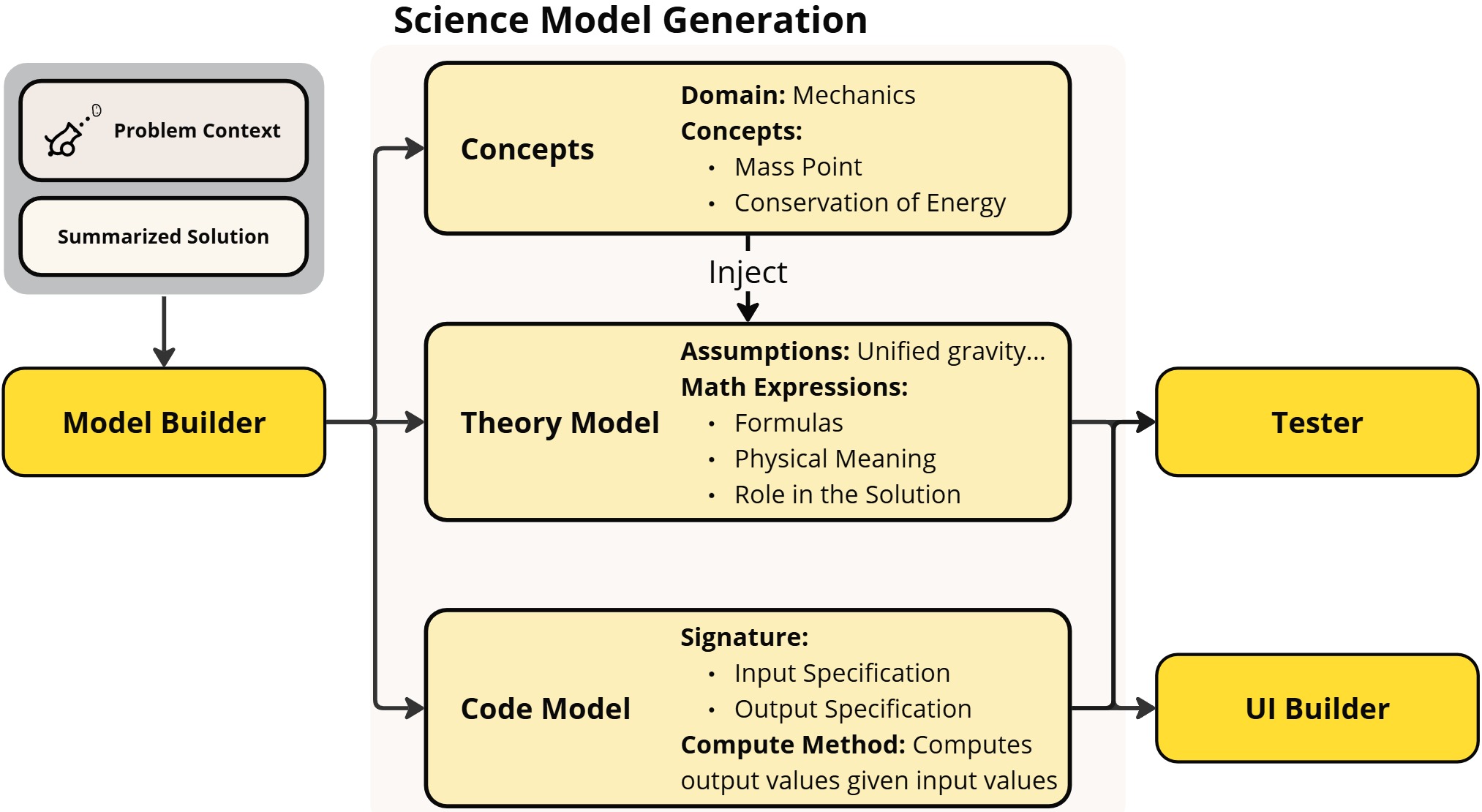}
    \caption{The model builder generates science models from summarized solutions, giving rise to interpretable reasoning}
    \label{fig:enter-label}
\end{figure*}

The science model and its interfaces are only practical for human scientists when they are faithful to the original reasoning result. To ensure that the science model and UI accurately reflect the original reasoning, we evaluate the consistency of our module using a subset of problems from the SciBench dataset. This subset contains problems from three textbooks: Fundamentals of Physics\cite{halliday2013fundamentals}, Statistical Thermodynamics\cite{engel2010statistical}, and Classical Dynamics of Particles and Systems\cite{thornton2021classical}. For a meaningful assessment, we carefully selected 50 problems, excluding those that involve only basic computations or contain incorrect reference solutions.

We evaluate consistency on two key dimensions:
\begin{itemize}
    \item \textbf{Numerical Consistency:} The science model should yield numerical results that agree with the original reasoning output.
    \item \textbf{Theoretical Consistency:} The constructed model should be physically coherent and correctly reflect the solution’s underlying principles.
\end{itemize}

Numerical consistency is verified via program execution, while theoretical consistency is assessed by a ChatGPT-4o model acting as a grader. The grader classifies each solution into three categories: highly consistent, moderately consistent, or inconsistent. We evaluate our model builders using two different underlying LLMs for agents: ChatGPT-4o and ChatGPT-4o-mini.

Table~\ref{tab:consistency} summarizes the numerical consistency of the base models. Although most solutions are consistent, discrepancies—stemming from reasoning failures or incorrect numerical outcomes—provide valuable feedback for further investigation by human experts.

Table~\ref{tab:consistency_levels} presents the theoretical consistency results. ChatGPT-4o demonstrates a higher degree of theoretical consistency, with no instances classified as completely inconsistent. This suggests that LLMs can effectively structure physics problems into theory models for interpretability.

\begin{table}
  \centering
  \begin{tabular}{lcc}
    \hline
    \textbf{Model} & \textbf{Cons.} & \textbf{Incons.} \\
    \hline
    ChatGPT-4o-mini & 47 & 3 \\
    ChatGPT-4o      & 46 & 4 \\
    \hline
  \end{tabular}
  \caption{Numerical Consistency of Different Base Models.}
  \label{tab:consistency}
\end{table}

\begin{table}
  \centering
  \begin{tabular}{lccc}
    \hline
    \textbf{Model} & \textbf{High} & \textbf{Mod.} & \textbf{Incons.} \\
    \hline
    ChatGPT-4o-mini & 43 & 3 & 4 \\
    ChatGPT-4o      & 47 & 3 & 0 \\
    \hline
  \end{tabular}
  \caption{Theoretical Consistency of Different Base Models.}
  \label{tab:consistency_levels}
\end{table}

\subsection{LLM Auxiliary Tester}
\begin{figure*}[ht]
    \centering
    \includegraphics[width=0.9\linewidth]{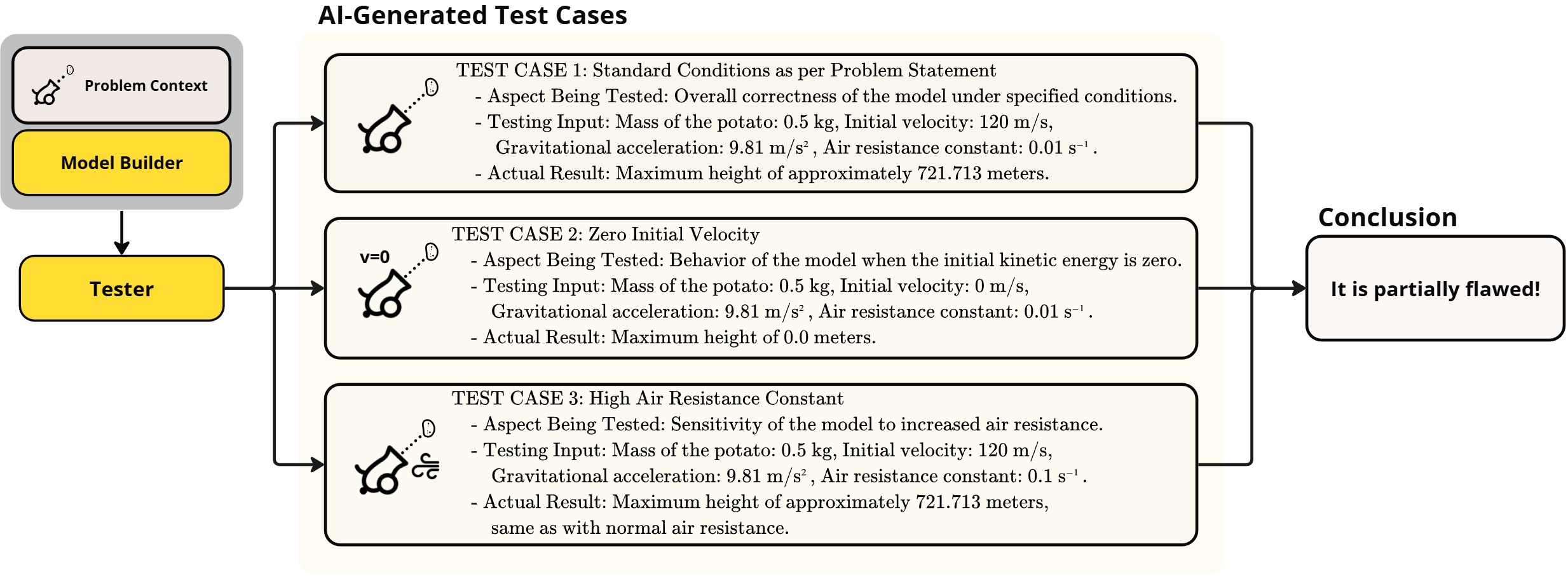}
    \caption{The Tester agent automatically generates test cases based on human-like reasoning principles}
    \label{fig:testcases}
\end{figure*}
In addition to generating solutions, the auxiliary tester enhances validation by automatically generating diverse test cases and analyzing their outcomes using the science model. Although LLM-generated test cases are common in software engineering \cite{tufano2020unit, li2022competition}, they also provide valuable insights when applied to science models. Our experiments show that LLMs naturally adopt human-like reasoning in test case generation, such as evaluating extreme scenarios. This enables them to provide more informative feedback beyond the science model and the interactive UI.

As depicted in Fig. \ref{fig:testcases}, the tester agent uncovers partial flaws in the model by exploring various input conditions, by reconsidering the original input and tuning the initial velocities and the air resistance constant. The tester agent's conclusion well aligns with the ground truth that the solution was indeed incorrect due to an erroneous underlying science model.

\subsection{Interactive UI}
\begin{figure*}[h]
    \centering
    \includegraphics[width=0.9\linewidth]{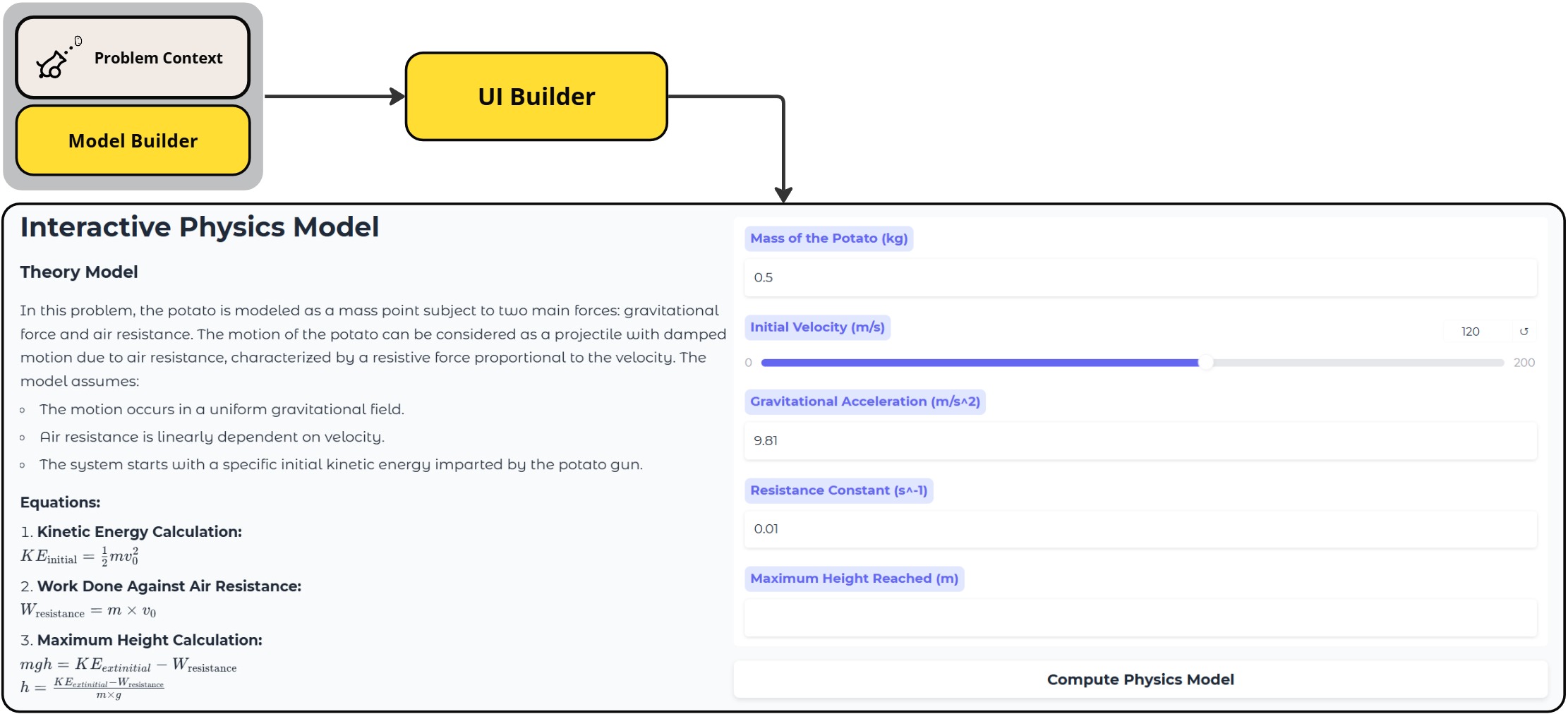}
    \caption{The interactive user interface enables intuitive feedback for human scientists}
    \label{fig:ui}
\end{figure*}

Inspired by previous work on enabling LLMs to generate user interfaces through coding \cite{wu2024uicoderfinetuninglargelanguage}, we introduce an interactive interface built using Gradio \cite{abid2019gradio}. The UI Builder agent converts the code model from the previous stage into an interactive interface, significantly reducing the effort required for validation, as shown in Fig. \ref{fig:ui}. This interface allows human scientists to develop intuition about the underlying science model. Similar to the code model, the UI Builder agent follows a predefined template to ensure stability and consistency. For our experiment, the UI Builder is prompted with a predefined Gradio \cite{abid2019gradio} template as the starting point for UI.

\section{Conclusion and Future Work}
In this work, we have presented a novel multi-agent LLM physicists framework with an interpretation module that enhances the interpretability and verifiability of LLM-generated physics reasoning. By leveraging a multi-agent system including a summarizer, theory model builder, coding model builder, visualization builder, and auxiliary tester, we can transform complex LLM outputs into structured, transparent science models. Our case study on a SciBench mechanics problem demonstrated that this approach not only streamlines the reasoning process, but also empowers scientists to inspect, validate, and refine AI-generated solutions with ease. This integration of human-like test case generation and interactive validation bridges the gap between automated reasoning and human scientific intuition, marking a significant step toward more reliable AI-augmented reasoning.

Our future work will focus on extending our framework to encompass a broader range of physics domains and even other scientific fields. We aim to further refine each agent's capabilities, enhance the interactive elements of the UI, and integrate more sophisticated feedback loops between human experts and the system. Additional research will investigate scalability, the handling of increasingly complex models, and the integration of advanced techniques such as real-time interactive debugging and deeper reasoning transparency. These efforts are expected to foster better AI-Scientist understanding, ultimately paving the way for more trustworthy and effective AI-augmented reasoning.

\bibliography{reference}

\begin{thebibliography}{20}
\providecommand{\natexlab}[1]{#1}
\providecommand{\url}[1]{\texttt{#1}}
\expandafter\ifx\csname urlstyle\endcsname\relax
  \providecommand{\doi}[1]{doi: #1}\else
  \providecommand{\doi}{doi: \begingroup \urlstyle{rm}\Url}\fi

\bibitem[Abid et~al.(2019)Abid, Abdalla, Abid, Khan, Alfozan, and Zou]{abid2019gradio}
Abid, A., Abdalla, A., Abid, A., Khan, D., Alfozan, A., and Zou, J.
\newblock Gradio: Hassle-free sharing and testing of ml models in the wild.
\newblock \emph{arXiv preprint arXiv:1906.02569}, 2019.

\bibitem[Achiam et~al.(2023)Achiam, Adler, Agarwal, Ahmad, Akkaya, Aleman, Almeida, Altenschmidt, Altman, Anadkat, et~al.]{achiam2023gpt}
Achiam, J., Adler, S., Agarwal, S., Ahmad, L., Akkaya, I., Aleman, F.~L., Almeida, D., Altenschmidt, J., Altman, S., Anadkat, S., et~al.
\newblock Gpt-4 technical report.
\newblock \emph{arXiv preprint arXiv:2303.08774}, 2023.

\bibitem[Anand et~al.(2024)Anand, Prasad, Kirtani, Nair, Gupta, Garg, Gautam, Buldeo, and Shah]{anand2024enhancing}
Anand, A., Prasad, K., Kirtani, C., Nair, A.~R., Gupta, M., Garg, S., Gautam, A., Buldeo, S., and Shah, R.~R.
\newblock Enhancing llms for physics problem-solving using reinforcement learning with human-ai feedback.
\newblock \emph{arXiv preprint arXiv:2412.06827}, 2024.

\bibitem[Cecchi \& Babkin(2024)Cecchi and Babkin]{cecchi-babkin-2024-reportgpt}
Cecchi, L. and Babkin, P.
\newblock {R}eport{GPT}: Human-in-the-loop verifiable table-to-text generation.
\newblock In Dernoncourt, F., Preo{\c{t}}iuc-Pietro, D., and Shimorina, A. (eds.), \emph{Proceedings of the 2024 Conference on Empirical Methods in Natural Language Processing: Industry Track}, pp.\  529--537, Miami, Florida, US, November 2024. Association for Computational Linguistics.
\newblock \doi{10.18653/v1/2024.emnlp-industry.39}.
\newblock URL \url{https://aclanthology.org/2024.emnlp-industry.39/}.

\bibitem[Ding et~al.(2023)Ding, Cen, and Wei]{ding2023using}
Ding, J., Cen, Y., and Wei, X.
\newblock Using large language model to solve and explain physics word problems approaching human level.
\newblock \emph{arXiv preprint arXiv:2309.08182}, 2023.

\bibitem[Engel \& Reid(2010)Engel and Reid]{engel2010statistical}
Engel, T. and Reid, P.
\newblock \emph{Statistical\^{} Thermodynamics, t Kinetics}.
\newblock Prentice Hall, New York, 2010.

\bibitem[Halliday et~al.(2013)Halliday, Resnick, and Walker]{halliday2013fundamentals}
Halliday, D., Resnick, R., and Walker, J.
\newblock \emph{Fundamentals of physics}.
\newblock John Wiley \& Sons, 2013.

\bibitem[Hennigen et~al.(2023)Hennigen, Shen, Nrusimha, Gapp, Sontag, and Kim]{hennigen2023towards}
Hennigen, L.~T., Shen, S., Nrusimha, A., Gapp, B., Sontag, D., and Kim, Y.
\newblock Towards verifiable text generation with symbolic references.
\newblock \emph{arXiv preprint arXiv:2311.09188}, 2023.

\bibitem[Huang et~al.(2023)Huang, Mamidanna, Jangam, Zhou, and Gilpin]{huang2023largelanguagemodelsexplain}
Huang, S., Mamidanna, S., Jangam, S., Zhou, Y., and Gilpin, L.~H.
\newblock Can large language models explain themselves? a study of llm-generated self-explanations, 2023.
\newblock URL \url{https://arxiv.org/abs/2310.11207}.

\bibitem[Li et~al.(2024)Li, Hu, Sun, Hu, Ye, Shan, Chen, and Zhang]{li2024truthreader}
Li, D., Hu, X., Sun, Z., Hu, B., Ye, S., Shan, Z., Chen, Q., and Zhang, M.
\newblock Truthreader: Towards trustworthy document assistant chatbot with reliable attribution.
\newblock In \emph{Proceedings of the 2024 Conference on Empirical Methods in Natural Language Processing: System Demonstrations}, pp.\  89--100, 2024.

\bibitem[Li et~al.(2022)Li, Choi, Chung, Kushman, Schrittwieser, Leblond, Eccles, Keeling, Gimeno, Dal~Lago, et~al.]{li2022competition}
Li, Y., Choi, D., Chung, J., Kushman, N., Schrittwieser, J., Leblond, R., Eccles, T., Keeling, J., Gimeno, F., Dal~Lago, A., et~al.
\newblock Competition-level code generation with alphacode.
\newblock \emph{Science}, 378\penalty0 (6624):\penalty0 1092--1097, 2022.

\bibitem[Pan et~al.(2024)Pan, Mudur, Taranto, Tikhanovskaya, Venugopalan, Bahri, Brenner, and Kim]{pan2024quantummanybodyphysicscalculations}
Pan, H., Mudur, N., Taranto, W., Tikhanovskaya, M., Venugopalan, S., Bahri, Y., Brenner, M.~P., and Kim, E.-A.
\newblock Quantum many-body physics calculations with large language models, 2024.
\newblock URL \url{https://arxiv.org/abs/2403.03154}.

\bibitem[Pang et~al.(2024)Pang, Hong, Zhou, Lv, Yang, Liang, Han, and Zhang]{pang2024physics}
Pang, X., Hong, R., Zhou, Z., Lv, F., Yang, X., Liang, Z., Han, B., and Zhang, C.
\newblock Physics reasoner: Knowledge-augmented reasoning for solving physics problems with large language models.
\newblock \emph{arXiv preprint arXiv:2412.13791}, 2024.

\bibitem[Shen et~al.(2024)Shen, Zhou, Chen, and Liu]{shen2024citekit}
Shen, J., Zhou, T., Chen, Y., and Liu, K.
\newblock Citekit: A modular toolkit for large language model citation generation.
\newblock \emph{arXiv preprint arXiv:2408.04662}, 2024.

\bibitem[Thornton \& Marion(2021)Thornton and Marion]{thornton2021classical}
Thornton, S.~T. and Marion, J.~B.
\newblock \emph{Classical Dynamics of Particles and Systems}.
\newblock Cengage Learning, Boston, 2021.

\bibitem[Tufano et~al.(2020)Tufano, Drain, Svyatkovskiy, Deng, and Sundaresan]{tufano2020unit}
Tufano, M., Drain, D., Svyatkovskiy, A., Deng, S.~K., and Sundaresan, N.
\newblock Unit test case generation with transformers and focal context.
\newblock \emph{arXiv preprint arXiv:2009.05617}, 2020.

\bibitem[Wang et~al.(2023{\natexlab{a}})Wang, Hu, Lu, Zhu, Zhang, Subramaniam, Loomba, Zhang, Sun, and Wang]{wang2023scibench}
Wang, X., Hu, Z., Lu, P., Zhu, Y., Zhang, J., Subramaniam, S., Loomba, A.~R., Zhang, S., Sun, Y., and Wang, W.
\newblock Scibench: Evaluating college-level scientific problem-solving abilities of large language models.
\newblock \emph{arXiv preprint arXiv:2307.10635}, 2023{\natexlab{a}}.

\bibitem[Wang et~al.(2023{\natexlab{b}})Wang, Duan, Fox, and Srinivasa]{wang2023newton}
Wang, Y.~R., Duan, J., Fox, D., and Srinivasa, S.
\newblock Newton: Are large language models capable of physical reasoning?
\newblock \emph{arXiv preprint arXiv:2310.07018}, 2023{\natexlab{b}}.

\bibitem[Wu et~al.(2024)Wu, Schoop, Leung, Barik, Bigham, and Nichols]{wu2024uicoderfinetuninglargelanguage}
Wu, J., Schoop, E., Leung, A., Barik, T., Bigham, J.~P., and Nichols, J.
\newblock Uicoder: Finetuning large language models to generate user interface code through automated feedback, 2024.
\newblock URL \url{https://arxiv.org/abs/2406.07739}.

\bibitem[Zhang et~al.(2024)Zhang, Chen, Jin, Wang, Ji, Wang, and Han]{zhang2024comprehensive}
Zhang, Y., Chen, X., Jin, B., Wang, S., Ji, S., Wang, W., and Han, J.
\newblock A comprehensive survey of scientific large language models and their applications in scientific discovery.
\newblock \emph{arXiv preprint arXiv:2406.10833}, 2024.

\end{thebibliography}
\bibliographystyle{icml2025}

\appendix
\newpage
\onecolumn
\tcbset{
  simplebox/.style={
    colframe=black,
    colback=gray!20,
    boxrule=0.5pt,
    arc=0mm,
    outer arc=0mm,
    left=6pt,
    right=6pt,
    breakable
  }
}

\lstset{
  basicstyle=\ttfamily\small,
  backgroundcolor=\color{gray!20},
  frame=single,
  rulecolor=\color{black},
  breaklines=true,
  captionpos=b,
  xleftmargin=6pt,
  xrightmargin=6pt,
  numbers=none,
}

\section{Completion of Case Study in Section 4}
\subsection{Theory Model}\label{appendix:case-study-theory}
\begin{tcolorbox}[simplebox]
In this problem, the potato is modeled as a mass point subject to two main forces: gravitational force and air resistance. The motion is treated as a projectile with damped motion (air resistance proportional to velocity). The assumptions are:
\begin{itemize}
    \item Uniform gravitational field.
    \item Air resistance is linearly dependent on velocity.
    \item The system starts with a defined initial kinetic energy from the potato gun.
\end{itemize}

\textbf{Equations and Explanations:}

\begin{enumerate}
    \item \textbf{Kinetic Energy Calculation:}
    \begin{itemize}
        \item \textbf{Meaning:} Represents the energy imparted to the potato.
        \item \textbf{Equation:} \(\displaystyle KE_{\text{initial}} = \frac{1}{2} m v_0^2\)
    \end{itemize}
    
    \item \textbf{Work Done Against Air Resistance:}
    \begin{itemize}
        \item \textbf{Meaning:} Energy lost as the potato overcomes air resistance.
        \item \textbf{Equation:} \(\displaystyle W_{\text{resistance}} = m \times v_0\)
    \end{itemize}
    
    \item \textbf{Maximum Height Calculation:}
    \begin{itemize}
        \item \textbf{Meaning:} Kinetic energy converts into gravitational potential energy.
        \item \textbf{Equation:} 
        \[
        h = \frac{KE_{\text{initial}} - W_{\text{resistance}}}{m \times g}
        \]
    \end{itemize}
\end{enumerate}
\end{tcolorbox}
\newpage
\subsection{Code Model}\label{appendix:case-study-code}
\begin{tcolorbox}[simplebox]
\begin{lstlisting}
import math

class PhysicsModel:
    def __init__(self):
        self.inputs = {
            "mass": "The mass of the potato in kg",
            "initial_velocity": "The initial velocity in m/s",
            "gravitational_acceleration": "Acceleration due to gravity in m/s^2",
            "resistance_constant": "Resistance constant k in s^-1"
        }
        self.outputs = {
            "maximum_height": "The maximum height reached in meters"
        }

    def compute(self, **kwargs):
        m = kwargs.get("mass", 0.5)
        v0 = kwargs.get("initial_velocity", 120)
        g = kwargs.get("gravitational_acceleration", 9.81)
        k = kwargs.get("resistance_constant", 0.01)
        
        # Calculate initial kinetic energy
        KE_initial = 0.5 * m * v0**2
        
        # Calculate work done against air resistance (simplified model)
        W_resistance = m * v0
        
        # Calculate maximum height
        h = (KE_initial - W_resistance) / (m * g)
        return {"maximum_height": h}

# Example usage
physics_model = PhysicsModel()
inputs = {
    "mass": 0.5,
    "initial_velocity": 120,
    "gravitational_acceleration": 9.81,
    "resistance_constant": 0.01
}
print(physics_model.compute(**inputs))
\end{lstlisting}
\end{tcolorbox}
\newpage
\section{Demo of an Electrodynamic Problem}
\subsection{Problem Context and Direct Solution}
\begin{tcolorbox}[simplebox]
\textbf{Problem Statement:} 
In a spherical metal shell of radius $R$, an electron is shot from the center directly toward a tiny hole in the shell, through which it escapes. The shell is negatively charged with a surface charge density (charge per unit area) of $6.90 \times 10^{-13} \mathrm{C} / \mathrm{m}^2$. What is the magnitude of the electron's acceleration when it reaches radial distances $r=0.500 R$? 

\textbf{Direct Solution:}
The magnitude of the electron's acceleration when it reaches a radial distance \( r = 0.500 R \) is approximately \( 1.371 \times 10^{10} \). The answer is therefore \(\boxed{13705705091.187}\).
\end{tcolorbox}

\subsection{Summarized Solution}
\begin{tcolorbox}[simplebox]
Therefore, the solution to the problem is:

\begin{enumerate}
    \item \textbf{Identify Constants:}
    \begin{itemize}
        \item Surface charge density, \(\sigma = 6.90 \times 10^{-13} \, \text{C/m}^2\)
        \item Charge of an electron, \(e = 1.602 \times 10^{-19} \, \text{C}\)
        \item Mass of an electron, \(m_e = 9.109 \times 10^{-31} \, \text{kg}\)
        \item Permittivity of free space, \(\epsilon_0 = 8.854 \times 10^{-12} \, \text{C}^2/(\text{N} \cdot \text{m}^2)\)
    \end{itemize}
    
    \item \textbf{Calculate Electric Field:}
    \begin{itemize}
        \item The electric field inside the shell is given by \(E = \sigma / \epsilon_0\).
    \end{itemize}
    
    \item \textbf{Calculate Force on the Electron:}
    \begin{itemize}
        \item The force experienced by the electron is \(F = e \cdot E\).
    \end{itemize}
    
    \item \textbf{Calculate Acceleration:}
    \begin{itemize}
        \item The acceleration of the electron is \(a = F / m_e\).
    \end{itemize}
    
    \item \textbf{Output the Result:}
    \begin{itemize}
        \item The calculated acceleration is approximately \(1.371 \times 10^{10} \, \text{m/s}^2\).
    \end{itemize}
\end{enumerate}

The code used to solve the problem is:

\begin{lstlisting}
# Constants
sigma = 6.90e-13  # Surface charge density in C/m^2
e = 1.602e-19  # Charge of electron in C
m_e = 9.109e-31  # Mass of electron in kg
epsilon_0 = 8.854e-12  # Permittivity of free space in C^2/(Nm^2)

# Electric field inside the shell
E = sigma / epsilon_0

# Force on the electron
F = e * E

# Acceleration of the electron
a = F / m_e

# Output the result
print(round(a, 3))
\end{lstlisting}
\end{tcolorbox}

\newpage
\subsection{Theory Model}
\begin{tcolorbox}[simplebox]
In this problem, we are examining the motion of an electron within a charged spherical metal shell. The key physical principles involve electrostatics and conductor behavior. The model considers:

\begin{itemize}
    \item Behavior of electric fields inside a conducting shell
    \item Electrostatic forces on charged particles
    \item Motion of an electron under electromagnetic forces
\end{itemize}

\section*{Equations and Explanations}

\begin{enumerate}
    \item \textbf{Electric Field Inside Conductor:}
    \begin{itemize}
        \item \textbf{Physical Meaning:} According to Gauss's law, the electric field inside a conductor is zero.
        \item \textbf{Equation:}
        \[
        E_{\text{inside}} = 0
        \]
        \item \textbf{Role in Solution:} This principle determines that there should be no electric field affecting the electron inside the shell.
    \end{itemize}
    
    \item \textbf{Force on Electron:}
    \begin{itemize}
        \item \textbf{Physical Meaning:} With zero electric field, there is no electrostatic force on the electron.
        \item \textbf{Equation:}
        \[
        F = e \times E_{\text{inside}} = 0
        \]
        \item \textbf{Role in Solution:} This shows that the electron experiences no electrostatic force inside the shell.
    \end{itemize}
    
    \item \textbf{Electron Acceleration:}
    \begin{itemize}
        \item \textbf{Physical Meaning:} With no force acting on the electron, there is no acceleration.
        \item \textbf{Equation:}
        \[
        a = \frac{F}{m_e} = 0
        \]
        \item \textbf{Role in Solution:} This confirms that the electron moves with constant velocity inside the shell.
    \end{itemize}
\end{enumerate}
\end{tcolorbox}

\newpage
\subsection{Code Model}
\begin{tcolorbox}[simplebox]
\begin{lstlisting}
import math

class PhysicsModel:
    def __init__(self):
        self.inputs = {
            "sigma": "Surface charge density in C/m^2",
            "e": "Charge of electron in C",
            "m_e": "Mass of electron in kg",
            "epsilon_0": "Permittivity of free space in C^2/(N m^2)"
        }
        self.outputs = {
            "E": "Electric field inside the shell in N/C",
            "F": "Force on the electron in N",
            "a": "Acceleration of the electron in m/s^2"
        }

    def compute(self, **kwargs):
        # Inside a conductor, electric field is always zero
        E = 0
        
        # Force on electron (zero due to zero field)
        F = 0
        
        # Acceleration (zero due to zero force)
        a = 0
        
        return {"E": E, "F": F, "a": a}

# Example usage
model = PhysicsModel()
outputs = model.compute(
    sigma=6.90e-13,
    e=1.602e-19,
    m_e=9.109e-31,
    epsilon_0=8.854e-12
)
print(f"Acceleration: {outputs['a']} m/s^2")
\end{lstlisting}
\end{tcolorbox}

\newpage
\subsection{User Interface}
\begin{figure}[ht]
    \centering
    \includegraphics[width=0.6\linewidth]{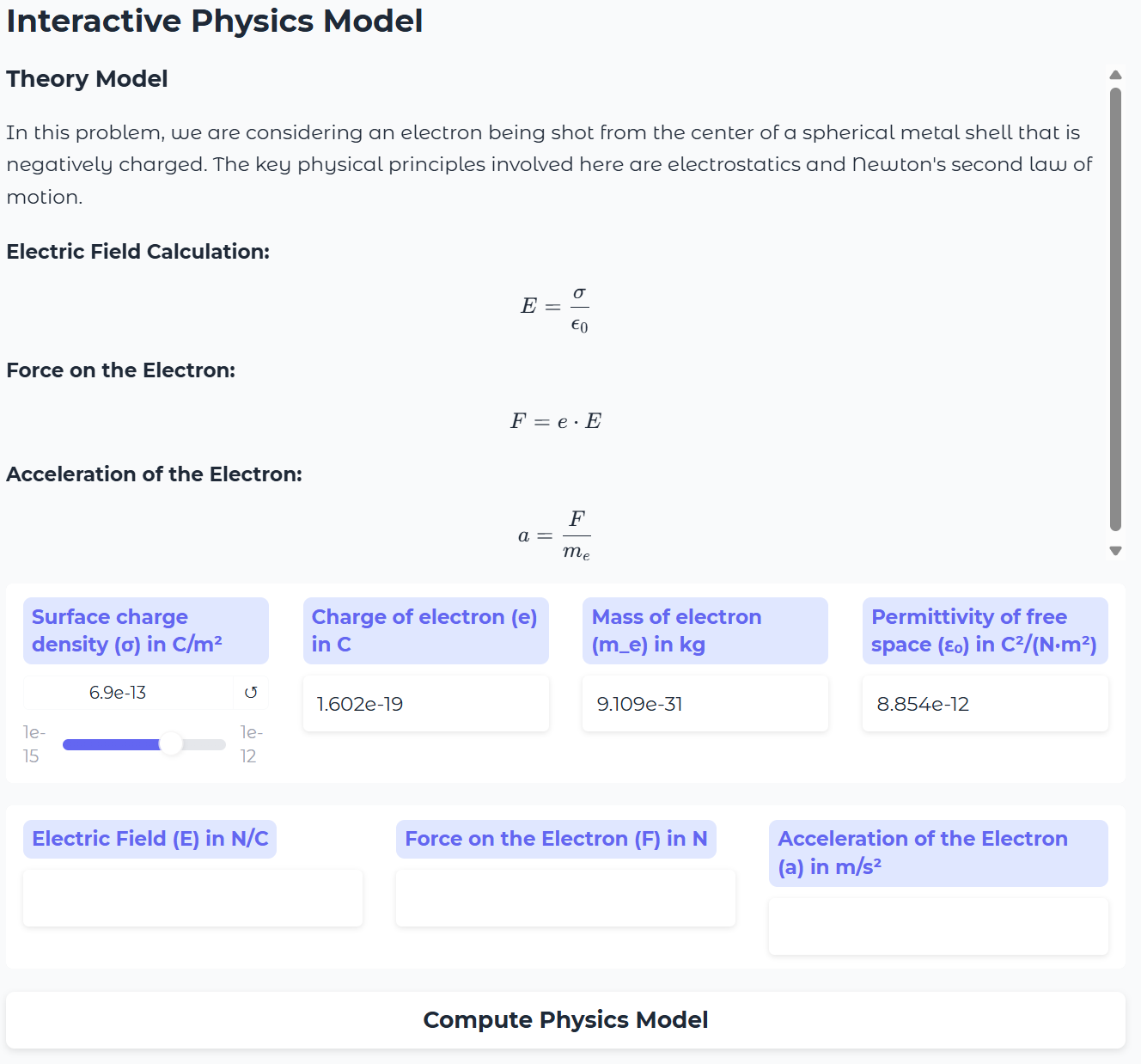}
    \caption{The interactive user interface for the Electrodynamic Problem}
    \label{fig:enter-label}
\end{figure}

\newpage
\subsection{Testing Results}

The tester successfully identifies that the model is flawed by discovering the discrepancies between the reasoning trajectory and the model.
\begin{tcolorbox}[simplebox]

\textbf{Test Case 1: Electric Field Inside Shell}
\begin{itemize}
    \item \textbf{Test Name:} Uniform Electric Field Inside the Shell
    \item \textbf{Testing Input:} $\sigma = 6.90 \times 10^{-13} \, \text{C/m}^2$, $\epsilon_0 = 8.854 \times 10^{-12} \, \text{C}^2/(\text{N} \cdot \text{m}^2)$
    \item \textbf{Expected Result:} $E = 0$ (According to Gauss's law)
    \item \textbf{Actual Result:} $E = \frac{\sigma}{\epsilon_0} \neq 0$
    \item \textbf{Finding:} FAILED - Model incorrectly assumes non-zero field
\end{itemize}

\textbf{Test Case 2: Force Calculation}
\begin{itemize}
    \item \textbf{Test Name:} Force Calculation on the Electron
    \item \textbf{Testing Input:} $e = 1.602 \times 10^{-19} \, \text{C}$
    \item \textbf{Expected Result:} $F = 0$ (Due to zero electric field)
    \item \textbf{Actual Result:} $F = e \cdot E \neq 0$
    \item \textbf{Finding:} FAILED - Incorrect force calculation
\end{itemize}

\textbf{Test Case 3: Acceleration Analysis}
\begin{itemize}
    \item \textbf{Test Name:} Acceleration Calculation Due to Electrostatic Force
    \item \textbf{Testing Input:} $m_e = 9.109 \times 10^{-31} \, \text{kg}$
    \item \textbf{Expected Result:} $a = 0$ (Due to zero force)
    \item \textbf{Actual Result:} $a = \frac{F}{m_e} \neq 0$
    \item \textbf{Finding:} FAILED - Incorrect acceleration calculation
\end{itemize}

\textbf{Final Assessment:}
\begin{itemize}
    \item \textbf{Determination:} FUNDAMENTALLY FLAWED
    \item \textbf{Confidence Level:} HIGH
    \item \textbf{Key Issues:}
    \begin{enumerate}
        \item Incorrect electric field calculation inside conductor
        \item Erroneous force computation
        \item Invalid acceleration derivation
    \end{enumerate}
\end{itemize}

\textbf{Recommended Corrections:}
\begin{enumerate}
    \item Apply Gauss's law correctly for conductors ($E = 0$ inside)
    \item Revise force calculations based on zero electric field
    \item Update acceleration computations accordingly
\end{enumerate}
\end{tcolorbox}
\newpage
\section{Demo of an Application Problem}
\subsection{Problem Context and Direct Solution}
\begin{tcolorbox}[simplebox]
\textbf{Problem Statement:} 
A skier weighing $90 \mathrm{~kg}$ starts from rest down a hill inclined at $17^\circ$. He skis $100 \mathrm{~m}$ down the hill and then coasts for $70 \mathrm{~m}$ along level snow until he stops. Find the coefficient of kinetic friction between the skis and the snow.

\textbf{Direct Solution:}
The coefficient of kinetic friction between the skis and the snow is approximately $0.177$. The answer is therefore \(\boxed{0.177}\).
\end{tcolorbox}

\newpage
\subsection{Summarized Solution}
\begin{tcolorbox}[simplebox]
The solution involves:
\begin{enumerate}
    \item \textbf{Energy Conservation:}
    \begin{itemize}
        \item Initial gravitational potential energy converts to kinetic energy and work against friction
        \item Final kinetic energy is zero when the skier stops
    \end{itemize}

    \item \textbf{Forces Analysis:}
    \begin{itemize}
        \item Gravitational force component: $F_{\parallel} = mg \sin(\theta)$
        \item Frictional force: $F_{\text{friction}} = \mu mg \cos(\theta)$
    \end{itemize}

    \item \textbf{Work-Energy Balance:}
    \begin{itemize}
        \item $mgh = F_{\text{friction}} \times (d_1 + d_2)$
        \item Where $h = d_1 \sin(\theta)$
    \end{itemize}

    \item \textbf{Final Equation:}
    \[
    \mu = \frac{d_1 \sin(\theta)}{d_2 + d_1 \cos(\theta)}
    \]
\end{enumerate}

The numerical solution was computed using Python:
\begin{lstlisting}[language=Python]
import math

g = 9.81  # acceleration due to gravity in m/s^2
d1 = 100  # distance down the hill in meters
d2 = 70   # distance along level snow in meters
theta = 17  # angle of incline in degrees

# Convert angle to radians
theta_rad = math.radians(theta)

# Calculate coefficient of friction
mu = (d1 * math.sin(theta_rad)) / (d2 + d1 * math.cos(theta_rad))
print(round(mu, 3))  # Result: 0.177
\end{lstlisting}
\end{tcolorbox}

\subsection{Theory Model}
\begin{tcolorbox}[simplebox]
The model is based on energy conservation and the work-energy theorem, applied to a skier descending an inclined plane and coasting on a level surface. Gravitational potential energy converts into kinetic energy and work against friction. The skier is modeled as a rigid body with constant mass, influenced only by gravity and friction. The friction on both the incline and level snow is characterized by a constant coefficient of kinetic friction, \(\mu\), which is to be determined.

\textbf{Key Assumptions:}
\begin{enumerate}
    \item The skier starts from rest (initial kinetic energy is zero).
    \item Friction is the only non-conservative force opposing the motion.
    \item Frictional force is proportional to the normal force, with a coefficient \(\mu\).
    \item The incline is uniform, and the transition to level snow involves no energy loss except for friction.
    \item Air resistance and other dissipative forces are negligible.
\end{enumerate}

\textbf{Equations and Explanations:}

\begin{enumerate}
    \item \textbf{Gravitational Force Parallel to the Incline:}
    \begin{itemize}
        \item \textbf{Physical Meaning:} This force accelerates the skier down the incline.
        \item \textbf{Equation:} 
        \[
        F_{\parallel} = mg \sin(\theta)
        \]
        \item \textbf{Role:} Provides the energy that is converted into kinetic energy and work against friction.
    \end{itemize}

    \item \textbf{Frictional Force on the Incline:}
    \begin{itemize}
        \item \textbf{Physical Meaning:} This force opposes the skier’s motion, proportional to the normal force.
        \item \textbf{Equation:} 
        \[
        F_{\text{friction, incline}} = \mu mg \cos(\theta)
        \]
        \item \textbf{Role:} Accounts for energy lost to friction as the skier descends.
    \end{itemize}

    \item \textbf{Energy Conservation and Work-Energy Principle:}
    \begin{itemize}
        \item \textbf{Physical Meaning:} Total mechanical energy loss equals work done by friction.
        \item \textbf{Equation:}
        \[
        mgh = \frac{1}{2} mv^2 + F_{\text{friction}} \times (d_1 + d_2)
        \]
        \item \textbf{Role:} Relates potential energy to energy dissipated by friction, enabling calculation of \(\mu\).
    \end{itemize}

    \item \textbf{Expression for Height:}
    \begin{itemize}
        \item \textbf{Physical Meaning:} Height \(h\) is the vertical displacement related to the initial gravitational potential energy.
        \item \textbf{Equation:}
        \[
        h = d_1 \sin(\theta)
        \]
        \item \textbf{Role:} Links the incline distance to potential energy in the energy conservation equation.
    \end{itemize}

    \item \textbf{Equation for Coefficient of Kinetic Friction \(\mu\):}
    \begin{itemize}
        \item \textbf{Physical Meaning:} Provides a direct relationship to calculate the coefficient \(\mu\).
        \item \textbf{Equation:}
        \[
        \mu = \frac{d_1 \sin(\theta)}{d_2 + d_1 \cos(\theta)}
        \]
        \item \textbf{Role:} Solving this equation determines \(\mu\) by equating gravitational energy conversion to energy dissipated by friction.
    \end{itemize}
\end{enumerate} 
\end{tcolorbox}

\newpage
\subsection{Code Model}
\begin{tcolorbox}[simplebox]
\begin{lstlisting}[language=Python]
import math

class PhysicsModel:
    def __init__(self):
        self.inputs = {
            "g": "The acceleration due to gravity in m/s^2",
            "d1": "The distance down the hill in meters",
            "d2": "The distance along the level snow in meters",
            "theta": "The angle of incline in degrees"
        }
        self.outputs = {
            "mu": "The coefficient of kinetic friction"
        }

    def list_inputs(self):
        """
        List the inputs required for the physics model, along with their physics meaning
        """
        return list(self.inputs.keys())
    
    def list_outputs(self):
        """
        List the outputs of the physics model, along with their physics meaning
        """
        return list(self.outputs.keys())
    
    def compute(self, **kwargs):
        """
        Compute the output of the physics model given the inputs

        Args:
            **kwargs: The inputs to the physics model

        Returns:
            dict: The computed outputs of the physics model
        """
        g = kwargs.get("g", 9.81)
        d1 = kwargs.get("d1", 100)
        d2 = kwargs.get("d2", 70)
        theta = kwargs.get("theta", 17)

        # Convert angle to radians for calculation
        theta_rad = math.radians(theta)

        # Calculate the coefficient of kinetic friction
        mu = (d1 * math.sin(theta_rad)) / (d2 + d1 * math.cos(theta_rad))

        # Format the answer to three decimal places
        mu_rounded = round(mu, 3)
        
        return {"mu": mu_rounded}

# Example usage
model = PhysicsModel()
inputs = {
    "g": 9.81,
    "d1": 100,
    "d2": 70,
    "theta": 17
}
outputs = model.compute(**inputs)
print(outputs["mu"])
\end{lstlisting}
\end{tcolorbox}

\newpage
\subsection{User Interface}
\begin{figure}[ht]
    \centering
    \includegraphics[width=0.6\linewidth]{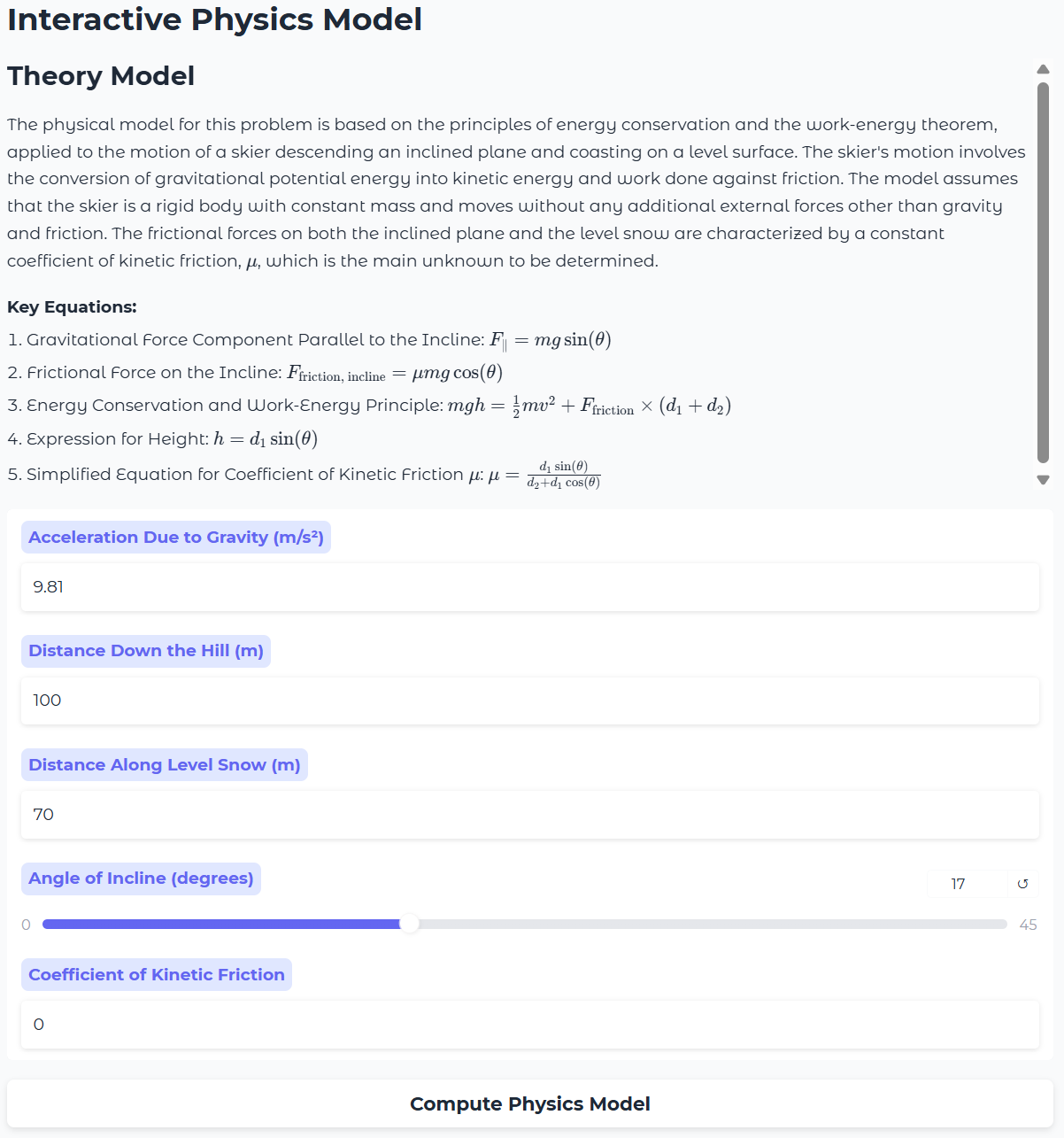}
    \caption{The interactive user interface for the Application Problem}
    \label{fig:enter-label}
\end{figure}

\newpage
\subsection{Testing Results}
The tester successfully confirms that the model and the reasoning are correct.
\begin{tcolorbox}[simplebox]
\textbf{Test Cases:}
\begin{enumerate}
    \item \textbf{Standard Case}
    \begin{itemize}
        \item Input: $d_1 = 100\text{ m}, d_2 = 70\text{ m}, \theta = 17^\circ$
        \item Result: $\mu = 0.177$
        \item Status: PASSED
    \end{itemize}

    \item \textbf{Steeper Angle}
    \begin{itemize}
        \item Input: $d_1 = 100\text{ m}, d_2 = 70\text{ m}, \theta = 25^\circ$
        \item Result: $\mu = 0.263$
        \item Status: PASSED
    \end{itemize}

    \item \textbf{Level Surface Only}
    \begin{itemize}
        \item Input: $d_1 = 0\text{ m}, d_2 = 70\text{ m}, \theta = 17^\circ$
        \item Result: $\mu = 0.000$
        \item Status: PASSED
    \end{itemize}
\end{enumerate}

\textbf{Final Assessment:}
\begin{itemize}
    \item \textbf{Model Status:} VERIFIED
    \item \textbf{Confidence Level:} HIGH
    \item \textbf{Key Findings:}
    \begin{enumerate}
        \item Model correctly handles standard input parameters
        \item Results scale appropriately with angle changes
        \item Edge cases produce physically meaningful results
    \end{enumerate}
\end{itemize}
\end{tcolorbox}

\end{document}